\documentclass[runningheads]{llncs}
\pdfoutput=1 

\usepackage[T1]{fontenc}
\usepackage{graphicx}
\usepackage[bookmarks=false]{hyperref}
\usepackage{color}

\usepackage{epstopdf}
\usepackage{amsmath}

\usepackage{cite}
\usepackage{comment}
\usepackage{cleveref}
\Crefname{figure}{Fig.}{Figs.}
\Crefname{section}{section}{sections}
\usepackage{multirow}
\usepackage{subfigure}
\usepackage{bm}

\begin{document}
\title{Temporal Performance Prediction for Deep Convolutional Long Short-Term Memory Networks}
\titlerunning{Temporal Performance Prediction for Deep ConvLSTMs}
\author{Laura Fieback\inst{1}\orcidID{0009-0003-3456-6766} \and
Bidya Dash\inst{1}\orcidID{0000-0002-8043-6669} \and
Jakob Spiegelberg\inst{1}\orcidID{0000-0002-6550-0087} \and
Hanno Gottschalk\inst{2}\orcidID{0000-0003-2167-2028}}

\authorrunning{L. Fieback et al.}

\institute{Volkswagen AG, Berliner Ring 2, 38440 Wolfsburg, Germany\\
\email{\{laura.fieback,bidya.binayam.dash,jakob.spiegelberg\}@volkswagen.de} \and
Mathematical Modeling of Industrial Life Cycles, Institute of Mathematics, TU Berlin, Berlin, Germany\\
\email{gottschalk@math.tu-berlin.de}}

\maketitle              

\begin{abstract}
Quantifying predictive uncertainty of deep semantic segmentation networks is essential in safety-critical tasks. In applications like autonomous driving, where video data is available, convolutional long short-term memory networks are capable of not only providing semantic segmentations but also predicting the segmentations of the next timesteps. These models use cell states to broadcast information from previous data by taking a time series of inputs to predict one or even further steps into the future. We present a temporal postprocessing method which estimates the prediction performance of convolutional long short-term memory networks by either predicting the intersection over union of predicted and ground truth segments or classifying between intersection over union being equal to zero or greater than zero. To this end, we create temporal cell state-based input metrics per segment and investigate different models for the estimation of the predictive quality based on these metrics. We further study the influence of the number of considered cell states for the proposed metrics.

\end{abstract}

\section{Introduction}
Retrieving information from images is an important task for scene understanding. Semantic image segmentation is a common approach to gain knowledge about image content by assigning each pixel a label from a predefined label space using neural networks. In safety-critical applications like autonomous driving \cite{Huang.2018} or medical diagnostics \cite{Wickstrm.2018}, information about the reliability of a prediction is indispensable for decision making. While most approaches to uncertainty quantification focus on a single frame only, temporal information is often available as in the case of video data. To leverage on this, we build on the meta classification and regression approach from \cite{Rottmann.2020} and \cite{Maag.2020}. The method introduced in \cite{Rottmann.2020} provides a framework to predict the performance of a segmentation network based on its softmax output, i.e., to predict the intersection over union $IoU$ (also known as Jaccard index \cite{Jaccard.1912}) per segment from metrics derived from its aggregated softmax outputs (meta regression) or classifying between $IoU=0$ and $IoU>0$ (meta classification). In \cite{Maag.2020}, the approach of \cite{Rottmann.2020} is extended to time series metrics using a light-weight tracking algorithm. In this work, we investigate temporal metrics retrieved from convolutional long short-term memory networks (ConvLSTMs). Long short-term memory networks (LSTMs) \cite{Hochreiter.1997} take time series as inputs to make predictions for future timesteps. Thus, the metrics presented in this work express uncertainties in single frames by taking account of temporal information from LSTM outputs. Moreover, we use the light-weight tracking algorithm from \cite{Maag.2020} to investigate the power of LSTM meta models. This is the first work that conducts meta classification and regression by considering LSTM-based temporal metrics and meta models. Note that our procedure requires a semantic segmentation LSTM network and a video stream of input data.

In our experiments, we use a ConvLSTM network \cite{Shi.2015} trained on the VIsual PERception (VIPER) dataset \cite{Richter.2017}. Our network takes a time series of semantic segmentations as input to predict the segmentation for the next timestep. We achieve classification accuracy of $96.15\%(\pm0.17\%)$ and $AUROC$ of $95.04\%$ $(\pm0.22\%)$. The best classification results using time series temporal metrics are obtained by our proposed LSTM meta model. For meta regression, we obtain $R^2$ values of $74.31\%(\pm0.33\%)$.

The remainder of this work is organized as follows. An overview over related work in the field of uncertainty quantification and object tracking is provided in \Cref{sec:relatedwork}. In \Cref{sec:metrics}, we introduce the temporal metrics for time-dynamic uncertainty quantification followed by the light-weight tracking algorithm in \Cref{sec:tracking}. In \Cref{sec:metatask} we describe the meta classification and regression method for time-dynamic performance prediction. Finally, we present our numerical results in \Cref{sec:results}.

\section{Related Work}\label{sec:relatedwork}
\subsection{Uncertainty Quantification}
Modern neural networks tend to be overconfident in their predictions \cite{Guo.2017, Minderer.2021}. Temperature scaling \cite{Guo.2017} and Dirichlet calibration \cite{Kull.2019} are scaling methods to calibrate the model's confidence estimates. Another common approach to quantify model uncertainty are Bayesian models \cite{MacKay.1992}. Different methods have been established to conduct Bayesian inference via variational approximations like \cite{Blundell.2015} and \cite{Duvenaud.2016}. In \cite{Huang.2018}, the sampling procedure is simulated based on temporal information in video data. Besides, Monte Carlo dropout \cite{Gal.2016} is widely used to approximate Bayesian neural networks. In \cite{Lakshminarayanan.2017}, deep ensembles are proposed to quantify predictive uncertainty based on the variance of the ensemble prediction. Other approaches like \cite{Riedlinger.2023} and \cite{Hornauer.2022} propose to model predictive uncertainty based on gradients. In \cite{Rottmann.2020}, a meta learning approach for semantic segmentation networks is introduced for false positive detection (meta classification) and performance prediction in terms of $IoU$ (meta regression). In \cite{Rottmann.2019} and \cite{Maag.2020}, this work is extended by adding resolution dependent uncertainty and temporal metrics, respectively. In \cite{Erdem.2004}, performance metrics for video object segmentation and tracking are introduced.

\subsection{Object Tracking}
Most works in the field of object tracking refer to the task of multi-object tracking, that is, tracking multiple objects in videos by means of bounding boxes \cite{Bergmann.2019, Peng.2020}. Tracking-by-detection \cite{Babenko.2009} is a common approach for this task, which separates objects from the background. The approaches in \cite{Wang.2019} and \cite{Belagiannis.2012} are based on segmentation and perform tracking using fully-convolutional Siamese networks and particle filter, respectively. Video panoptic segmentation \cite{Kim.2020} combines the task of semantic segmentation and object tracking at the same time. Recent works in this field \cite{Hurtado.2020, Kim.2020} propose end-to-end architectures to fulfill both tasks simultaneously. In \cite{Maag.2020}, a tracking algorithm is introduced which builds up on a semantic segmentation and matches segments of the same class based on their overlap in consecutive video frames.

\section{Segment-wise Dispersion and Temporal Metrics} \label{sec:metrics} 
We build metrics for the meta classification and regression task based on the output of our ConvLSTM video frame prediction model. The aim of our model is to predict the semantic segmentation of the next timestep given a video sequence of previous segmentations. Semantic segmentation can be viewed as a pixel-wise classification task, where each pixel $z$ of an input image $x$ is classified as a label $y \in C=\{y_1,\dots,y_c\}$ with $c$ possible output labels. The network's softmax output $f_z \left(y|x,w\right)$ can be interpreted as a probability distribution over the output labels $y \in C=\{y_1,\dots,y_c\}$ given the input image $x$ and the network weights $w$. The predicted class for a pixel $z$ is then given by the largest softmax value, i.e.,
\begin{equation}
	\hat{y}_z \left(x,w\right) = \underset{y \in C}{\mathrm{argmax}} f_z \left(y|x,w\right).
\end{equation}
The degree of randomness in a network's softmax output can be quantified using dispersion measures. Thus, we build metrics for the meta classification and regression task based on uncertainty heatmaps representing pixel-wise dispersion measure as proposed in \cite{Rottmann.2019}. We consider the entropy
\begin{equation}
	E_z\left( x,w\right) = - \frac{1}{\log\left( c \right)} \sum_{y \in C} f_z \left(y|x,w\right) \log f_z \left(y|x,w\right),
\end{equation}
the variation ratio
\begin{equation}
	V_z\left( x,w \right) = 1 - \max_{y \in C } f_z \left(y|x,w\right),
\end{equation}
as well as the probability margin
\begin{equation}
	M_z\left( x,w \right) = 1 - \max_{y \in C } f_z \left(y|x,w\right) + \max_{y \in C \setminus \hat{y}_z } f_z \left(y|x,w\right).
\end{equation}
Note that, for better comparison, these quantities have been normalized to the interval $[0,1]$. Let $\hat{S}_x = \{ \hat{y}_z \left( x,w \right) | z \in x \}$ denote the predicted semantic segmentation for an image $x$ and $\hat{\mathcal{K}}_x$ the set of all segments $k$ in $x$, i.e., the set of all connected components of pixels $z'$ with the same predicted class $\hat{y}_{z'} = c'$.

The segment-wise dispersion metrics based on the pixel-wise uncertainty heatmaps introduced above are defined as
\begin{equation}\label{eq5}
	\bar{D} = \frac{1}{S} \sum_{z \in k} D_z \left( x,w \right),
\end{equation}
where $D_z \in \{E_z , V_z , M_z\}$ and $S = | \{ z \in k \} |$ denotes the number of pixels contained in $k$, that is, the segment size. As proposed in \cite{Rottmann.2020}, we define inner dispersion metrics and boundary dispersion metrics, since we typically observe high values of $D_z$ for boundary pixels. To this end, let $k_{in} \subset k$ denote the set of all inner pixels of segment $k$, where a pixel $z \in k$ is called an inner pixel of $k$ if all eight neighboring pixels are an element of $k$, and let $k_{bd} = k \setminus k_{in}$ denote the set of boundary pixels of segment $k$. We obtain further segment-wise dispersion metrics by averaging the pixel-wise uncertainty heatmaps over all inner pixels and boundary pixels by analogy with equation \eqref{eq5} yielding the inner and boundary dispersion metrics $\bar{D}_{in}$ and $\bar{D}_{bd}$ as well as $S_{in}$ and $S_{bd}$. Based on these metrics, we obtain the respective relative metrics $\tilde{S} = S/S_{bd}$, $\tilde{S}_{in} = S_{in}/S_{bd}$, $\tilde{D} = \bar{D} \tilde{S}$ and $\tilde{D}_{in} = \bar{D}_{in} \tilde{S}_{in}$ with $D\in\{E,V,M\}$. Our set of metrics further contains the geometric center
\begin{equation}
	\bar{k} = \left(\bar{k}_1,\bar{k}_1\right) = \frac{1}{S} \sum_{z \in k} \left(z_1,z_1\right),
\end{equation}
where $z_1$ and $z_2$ are the vertical and horizontal coordinates of pixel $z$ as well as the mean class probabilities for each class $y \in C=\{y_1,\dots,y_c\}$,
\begin{equation}
	P\left(y|k\right) = \frac{1}{S} \sum_{z \in k} f_z \left(y|x,w\right).
\end{equation}
This results in the following set of metrics (see \cite{Maag.2020})
\begin{equation}
	\begin{split}
		U = & \{  \bar{D},  \bar{D}_{in},  \bar{D}_{bd}, \tilde{D}, \tilde{D}_{in} \ | \ D\in\{E,V,M\} \} \cup \{\bar{k}\} \\
		& \cup \{S, S_{in}, S_{bd}, \tilde{S}, \tilde{S}_{in}\} \cup \{P\left(y|k\right) | y=y_1,\dots,y_c\}.
	\end{split}
\end{equation}
\begin{figure}
	\includegraphics[width=\textwidth]{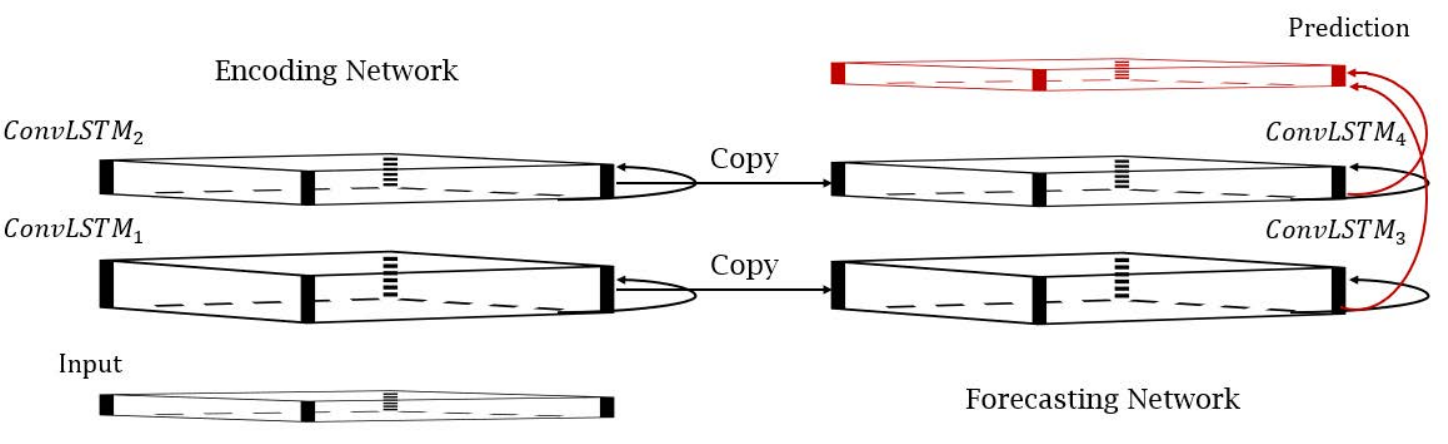}
	\caption{Depiction of a ConvLSTM block (from \protect\cite{Shi.2015}).} \label{fig:ConvLSTM}
\end{figure}
We use these metrics as a baseline in our tests and define additional metrics based on the cell states of our ConvLSTM video frame prediction model. Our model consists of $l=10$ ConvLSTM blocks using ten previous semantic segmentations $x_{t-i}, \ i=1, \dots, 10$ of a video to predict the semantic segmentation of the next video frame $\hat{x}_{t}$. Note that every ConvLSTM block itself consists of an encoding network and a forecasting network, where both networks consist of the same number of convolutional LSTM cells with shared hidden states and cell states (see \Cref{fig:ConvLSTM}). The shared hidden states and cell states between both networks are the same states, which are broadcasted to the next ConvLSTM block. In our model, the last convolutional LSTM cell of the forecasting network of each ConvLSTM block outputs states of the same height and width as the model's prediction with 64 features. Thus, for every ConvLSTM block, we focus on the cell state of the last convolutional LSTM cell and define the mean cell state $\bar{C}^{i} , \ i = 1, \dots, 10$ as the mean over the 64 features. Based on this, we build temporal heatmaps from the stability of the mean cell state $\bar{C}^{i}$ over $i$\break ConvLSTM blocks. To this end, we define the stability of cell state $j$ for an image $x$, a pixel $z$ and network weights $w$ as
\begin{equation}
	C^j_z \left(x,w\right) = |\bar{C}^{1}_z \left(x,w\right) - \bar{C}^{j+1}_z \left(x,w\right)|, \ j = 1, \dots, 9.
\end{equation}
As for the uncertainty heatmaps introduced above, we define segment-wise temporal metrics based on the temporal heatmaps as
\begin{equation}
	\bar{T} = \frac{1}{S} \sum_{z \in k} T_z \left( x,w \right),
\end{equation}
with $T_z \in \{C^j_z , \ j = 1, \dots, 9\}$. With the notation above, we define our proposed set of metrics for $m =1,\dots,9$ as 
\begin{equation}
	V_m = U \cup CS_m,
\end{equation}
where
\begin{equation}
	CS_m =  \{  \bar{T},  \bar{T}_{in},  \bar{T}_{bd}, \tilde{T}, \tilde{T}_{in} \ | \ T \in \{C^j , \ j = 1, \dots, m\}\}.
\end{equation}
Note that all of these metrics can be calculated from our model output without any knowledge of the ground truth.

\section{Tracking Algorithm} \label{sec:tracking}
For the investigation of LSTMs as meta models, we apply the tracking algorithm proposed in \cite{Maag.2020}. This algorithm builds on a video sequence of segmentations and performs tracking based on the overlap of segments of the same class in consecutive frames. It does not require additional training. Within this procedure, every segment is assigned a tracking id. To this end, let $\{x_1,\dots,x_{T}\}$ denote a sequence of $T$ consecutive semantic segmentations. The overlap of a segment $k$ with segment $j$ is defined as
\begin{equation}
	O_{j,k} = \frac{|\{  z \in k \} \cap \{ z \in j \}|}{|\{ z \in j \}|}.
\end{equation}
The algorithm is applied sequentially to each segmentation $x_t, \ t = 1,\dots,T$, where for each frame, the segments are prioritized based on their segment size. In detail, the algorithm consists of five steps starting with the largest segment $k^{S_{max}} \in \hat{\mathcal{K}}_{x_t}$ in each step. Once a segment $k \in \hat{\mathcal{K}}_{x_t}$ has been matched with a segment from a previous frame, it is ignored in the following steps. Matched segments receive the same tracking id. To this end, we denote a matched segment $k$ in $x_t$ as $k_t$.

\textbf{Step 1} matches segments of the same class in $x_t$ which are close to each other, i.e., with a distance less than a constant $c_{near}$, and thus, are regarded as one segment.

\textbf{Step 2} matches segments based on their geometric center. If a segment $k$ exists in two consecutive frames, i.e., $k \in  \hat{\mathcal{K}}_{x_{t-1}} \cap  \hat{\mathcal{K}}_{x_{t-2}}$, segment $k_{t-1}$ is shifted by $\left(\bar{k}_{t-1} - \bar{k}_{t-2}\right)$ and segments $j \in \hat{\mathcal{K}}_{x_{t}}$ are matched with the shifted segment $\hat{k}_{t}$, if the overlap $O_{j,\hat{k}_t}$ is higher than a constant $c_{over}$ or if the distance between the geometric centers $\bar{j}$ and $\bar{\hat{k}}_{t}$ is smaller than $c_{dist}$. If $k$ does not exist in two consecutive frames, segments $j \in \hat{\mathcal{K}}_{x_{t}}$ are matched based on the distance of the geometric centers $\bar{j}$ and $\bar{k}_{t-1}$.
		
\textbf{Step 3} matches segments in consecutive frames based on their overlap, i.e., segments $k \in \hat{\mathcal{K}}_{x_{t-1}}$ and $j \in \hat{\mathcal{K}}_{x_{t}}$ are matched if $O_{j,k}\geq c_{over}$.

\textbf{Step 4} accounts for flashing predicted segments due to occlusions or false predictions. It aims at matching segments that are more than one frame apart in temporal direction. To this end, a linear regression model is used to predict the geometric center of segment $k$ in $x_t$ if $k$ was matched in at least two of the last $lr$ segmentations $x_{t-1},\dots, x_{t-lr}$. Segments $j \in \hat{\mathcal{K}}_{x_{t}}$ are matched if the distance between the predicted geometric center $\hat{\bar{k}}_t$ and $\bar{j}$ is less than a constant $c_{lin}$.

\textbf{Step 5} assigns a new id to all segments $j \in \hat{\mathcal{K}}_{x_{t}}$, that have not yet been matched.

\section{IoU Prediction}\label{sec:metatask}
For the task of semantic segmentation, a common measure for predictive quality is the $IoU$. In our experiments, we use a slight modification proposed in \cite{Rottmann.2020}, the $IoU_{adj}$, which is less prone to fragmented objects. We perform segment-wise meta classification, that is, classifying between $IoU_{adj}=0$ and $IoU_{adj}>0$ as well as segment-wise meta regression, i.e., predicting $IoU_{adj}$ for each segment by means of the metrics defined in \Cref{sec:metrics}. Note that all of these metrics can be calculated from the ConvLSTM's output without any knowledge of the ground truth. We analyze the information gain induced by the temporal metrics for both, single frame metrics and time series metrics as proposed in \cite{Maag.2020}. Those time series metrics are based on the tracking algorithm introduced in \Cref{sec:tracking}. For each segment $k_t \in \hat{\mathcal{K}}_{x_{t}}$, we obtain single-frame based metrics $V^k_m=V^k_{m,t}$ according to \Cref{sec:metrics} as well as their history $V^k_{m,t-1},\dots , V^k_{m,t-T}$ due to tracking of segment $k$ over $T$ previous frames. In our experiments, we investigate the influence of metric histories for up to $T=10$ timesteps. In \cite{Maag.2020}, different models for the meta tasks were investigated. We choose the best performing models, i.e., the linear model (LR), the shallow neural network (NN) as well as the gradient boosting model (GB) for our experiments (for implementation details, see \cite{Maag.2020}). In addition, we investigate the performance of a shallow LSTM neural network (in the following referred to as LSTM) with $50$ neurons only for both meta tasks. The number of LSTM cells depends on the respective number of considered timesteps $T$ of the time series metrics.

\section{Numerical Results}\label{sec:results}
In this section, we investigate the properties of the temporal metrics defined in \Cref{sec:metrics}. We further investigate the influence of time series metrics as described in the previous section and consider different models for meta classification and regression. To this end, we train a ConvLSTM network with ten blocks, each of them built by five convolutional LSTM cells (see \Cref{fig:ConvLSTM}). We train our model on the synthetic VIPER dataset \cite{Richter.2017}. The dataset consists of more than $250.000$ frames all annotated with ground truth data with a resolution of $1920 \times 1080$ pixels per frame. Since the ground truth annotation has very fine labels, we apply the smoothing algorithm proposed in \cite{Rottmann.2023} to generate a coarse ground truth by blurring each class using a normalized box filter. Moreover, we resize the images to $256 \times 512$ pixels for computational reasons. The VIPER dataset contains $32$ different classes with $23$ proposed training ids. Out of these, we further cluster highly underrepresented classes to a misc class which results in a total of $17$ training classes. We train our ConvLSTM model on $19$ training folders which contain $30,168$ images in total and $8$ validation folders yielding a total of $7,021$ images. In our experiments, we compare two different models from our training procedure: The "strong model" (S) which was trained for $18$ epochs yielding a mean $IoU$ ($mIoU$) of $82.82\%$, as well as the "weak model" (W) which obtained an $mIoU$ of $79.45\%$ after $4$ epochs of training. We implement the tracking algorithm from \Cref{sec:tracking} with parameters $c_{near}=10$, $c_{over}=0.35$, $c_{dist}=100$ and $c_{lin}=50$.

For the meta tasks, we use $5$ validation folders, not yet used during the training procedure of the ConvLSTM model, which sum up to $3,464$ images. This results in a total of $46,587,336$ segments for the weak model (not yet matched over time) of which $110,739$ have non-empty interior. Out of these, $7,649$ segments have $IoU_{adj}=0$. For the strong model, we obtain $42,295,440$ segments, $113,286$ with non-empty interior of which $5,622$ segments have $IoU_{adj}=0$.\break The corresponding naive classification baseline discussed in \cite{Rottmann.2020} and \cite{Maag.2020} yields an accuracy of $93.09\%$ for the weak model and $95.04\%$ for the strong model. This baseline is obtained by random guessing, i.e., randomly assigning a probability to each segment and thresholding on it. The classification accuracy is the number of correct predictions divided by the total number of predictions made. The corresponding $AUROC$ value is $50\%$. This baseline is clearly outperformed. To this end note that, the stronger the ConvLSTM model, the higher the naive accuracy. We are able to improve the naive accuracy by further $1.63 pp$ for the weak model and $0.95 pp$ for the strong model.

In all our experiments, we average our results over ten randomly sampled train/val/test $(70\% / 10\% / 20\% )$ splits using $38,000$ segments in each split. In tables, the corresponding standard deviations are given in brackets, whereas, in figures, they are given by shades. All meta models considered yield an inference time for all $38,000$ segments together of less than one second. We measure the classification performance of our method in terms of classification accuracy ($ACC$) and Area Under Receiver Operating Characteristic ($AUROC$), which is obtained by varying the decision threshold between $IoU_{adj}=0$ and $IoU_{adj}>0$. For meta regression, we state the results in terms of the regression standard error $\sigma$ and the $R^2$ value.

\begin{table}[t]
	\caption{Results for meta classification and regression based on temporal metrics for different meta models and the entropy baseline for both, the weak (W) and the strong (S) model. The super script denotes the number of cell state metrics, where the best performance and in particular the given values are reached. The best results are highlighted.}\label{tab:temporalmetrics}
	\begin{center}
		\scriptsize
		\begin{tabular}{|| c || c || c | c | c | c ||} 
			\hline
			\multicolumn{6}{||c||}{Meta Classification $IoU_{adj}=0,>0$} \\
			\hline
			\multicolumn{6}{||c||}{	Entropy Baseline (W): $ACC = 93.40\%(\pm0.20\%)$ $AUROC = 81.63\%(\pm0.78\%)$} \\
			\multicolumn{6}{||c||}{	Entropy Baseline (S): $ACC = 95.27\%(\pm0.20\%)$ $AUROC = 80.45\%(\pm0.71\%)$} \\
			\hline
			\multicolumn{2}{||c}{} &
			\multicolumn{1}{c|}{GB} &
			\multicolumn{1}{|c|}{LR} &
			\multicolumn{1}{|c|}{LSTM} &
			\multicolumn{1}{|c||}{NN} \\
			\hline
			\multirow{2}*{$ACC$}
			& W & \bm{$94.72\%(\pm0.22\%)^7$} & $94.39\%(\pm0.16\%)^1$ & $94.01\%(\pm0.16\%)^6$ & $93.72\%(\pm0.22\%)^2$ \\
			& S &  \bm{$95.99\%(\pm0.17\%)^9$} & $95.65\%(\pm0.15\%)^9$ & $95.54\%(\pm0.22\%)^2$ & $95.35\%(\pm0.21\%)^6$ \\
			\hline
			\multirow{2}*{$AUROC$}
			& W &  \bm{$94.54\%(\pm0.44\%)^0$} & $93.69\%(\pm0.47\%)^2$ & $93.28\%(\pm0.53\%)^0$ & $92.85\%(\pm0.59\%)^0$ \\ 
			& S & \ \bm{$93.87\%(\pm0.43\%)^2$} & $92.57\%(\pm0.42\%)^9$ & $92.25\%(\pm0.44\%)^9$ & $91.87\%(\pm0.45\%)^9$ \\ 
			\hline
			\multicolumn{6}{||c||}{Meta Regression $IoU_{adj}$} \\
			\hline
			\multicolumn{6}{||c||}{Entropy Baseline (W): $\sigma = 0.227(\pm0.002)$ $R^2 = 42.80\%(\pm0.70\%)$} \\
			\multicolumn{6}{||c||}{Entropy Baseline (S): $\sigma = 0.225(\pm0.003)$ $R^2 = 38.58\%(\pm0.81\%)$} \\
			\hline
			\multicolumn{2}{||c}{} &
			\multicolumn{1}{c|}{GB} &
			\multicolumn{1}{|c|}{LR} &
			\multicolumn{1}{|c|}{LSTM} &
			\multicolumn{1}{|c||}{NN} \\
			\hline
			\multirow{2}*{$\sigma$}
			& W & \bm{$0.154\%(\pm0.002\%)^8$} & $0.175\%(\pm0.002\%)^0$ & $0.162\%(\pm0.001\%)^0$ & $0.155\%(\pm0.002\%)^8$ \\ 
			& S & $0.161\%(\pm0.001\%)^9$ & $0.175\%(\pm0.002\%)^0$ & $0.165\%(\pm0.001\%)^0$ & \ \bm{$0.160\%(\pm0.002\%)^9$} \\
			\hline
			\multirow{2}*{$R^2$}
			& W & \bm{$74.04\%(\pm0.52\%)^0$} & $66.85\%(\pm0.43\%)^9$ & $70.96\%(\pm0.47\%)^9$ & $73.57\%(\pm0.46\%)^0$ \\
			& S & $68.95\%(\pm0.61\%)^1$ & $63.33\%(\pm0.59\%)^8$ & $67.61\%(\pm0.43\%)^9$ & \ \bm{$69.19\%(\pm0.47\%)^3$} \\
			\hline
		\end{tabular}
	\end{center}
\end{table}

\begin{figure}[t]
	\begin{center}
		\subfigure[Weak Model $R^2$]{\includegraphics[width=.328\textwidth]{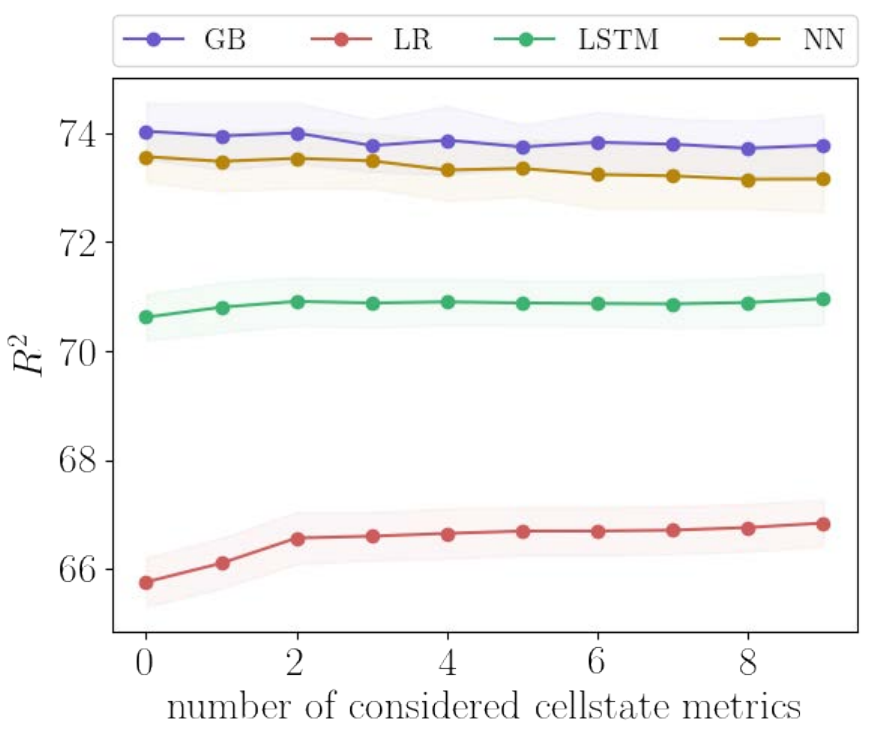}\label{fig:tempa}}
		\subfigure[Strong Model $R^2$]{\includegraphics[width=.328\textwidth]{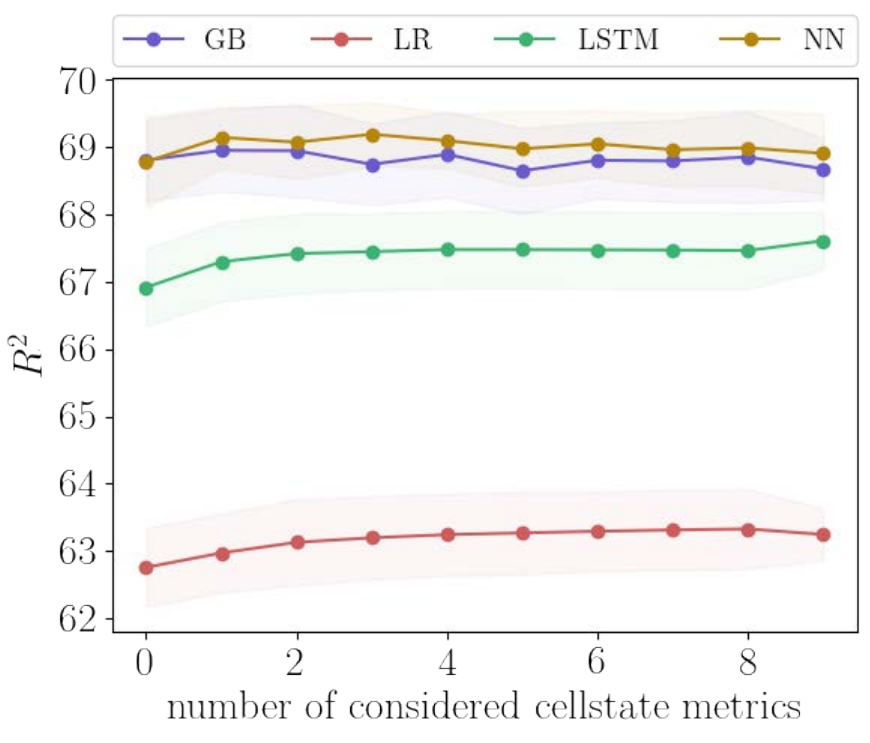}\label{fig:tempb}}
		\subfigure[Strong Model $ACC$]{\includegraphics[height=97pt,width=.328\textwidth]{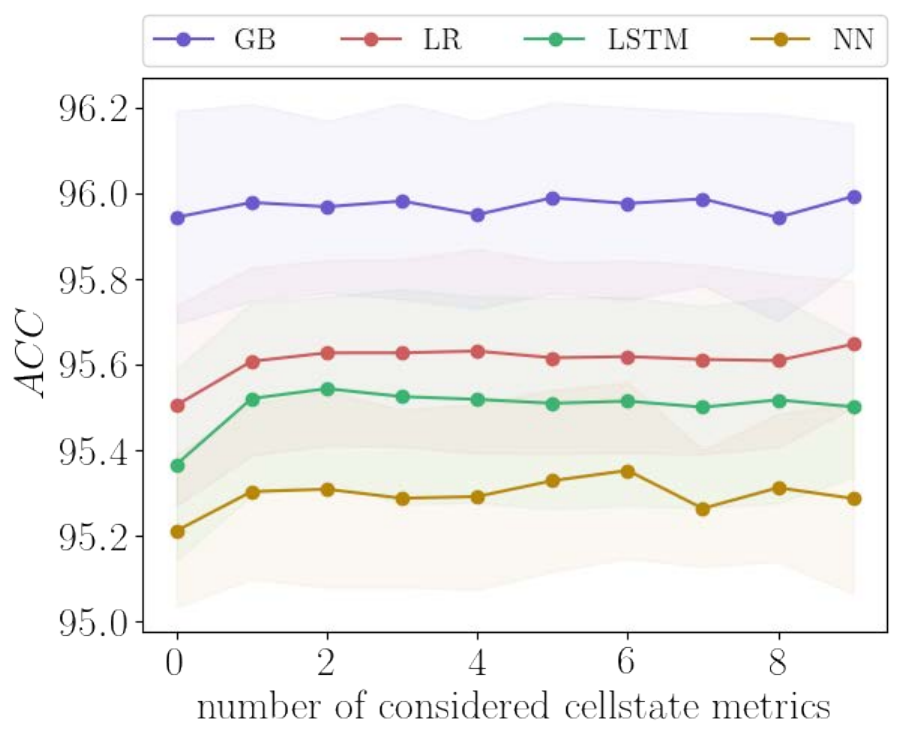}\label{fig:tempc}}
		\caption{A selection of results for meta classification in terms of $ACC$ and meta regression in terms of $R^2$ as functions of the number of considered cell state metrics. Meta regression via the weak model (a), meta regression via the strong model (b), meta classification via the strong model (c).}\label{fig:tempmetrics}
	\end{center}
\end{figure}

First, we investigate the influence of single-frame temporal metrics $V_m=V_{m,t}$ by considering the stability of cell states over $m\in\{1,\dots,9\}$ ConvLSTM blocks. \Cref{tab:temporalmetrics} shows the best results for different meta models. The super script denotes the number of considered cell state metrics, where the best performance and in particular the given values are reached. For the weak model, we achieve test $AUROC$ values of up to $94.54\%(\pm0.44\%)$ and classification accuracies of up to $94.72\%(\pm0.22\%)$. For the strong model, a test accuracy of $95.99\%(\pm0.17\%)$ is reached and $AUROC$ value up to $93.87\%(\pm0.43\%)$. As in \cite{Maag.2020}, GB performs best for meta classification. With regard to meta regression, we obtain $R^2$ values up to $74.04\%(\pm0.52\%)$ for the weak model and $69.19\%(\pm0.47\%)$ for the strong model. As a baseline, we consider the approach from \cite{Rottmann.2020}, i.e., the metric set $U_t$ without cell state metrics. In almost every experiment, best results are obtained when considering temporal metrics. In those cases, where the best results are obtained without temporal metrics, we observe vanishing differences between the respective performance metrics for temporal metrics (e.g., see $R^2$ values for GB and NN in \Cref{fig:tempa}). In \cite{Rottmann.2020}, the results are compared with the entropy as a single-metric baseline and with the naive baseline introduced above. For the entropy baseline (see \Cref{tab:temporalmetrics}), we use single-frame gradient boosting as suggested in \cite{Maag.2020}. Both baselines are clearly outperformed. In contrast to the results in \cite{Maag.2020}, the GB meta regression model does not outperform the neural network in all settings, even though it yields the best results in most of the experiments.

\Cref{fig:tempmetrics} shows the influence of temporal metrics with respect to $R^2$ value and classification accuracy. For the linear meta regression model based on the weak ConvLSTM (\Cref{fig:tempa}), we obtain $R^2$ values up to $66.85\%(\pm0.43\%)$ when taking account of all $m=9$ temporal metrics, whereas the baseline metrics $U_t$ ($0$ considered cell state metrics) only achieve averaged $R^2$ values of $65.77\%(\pm0.45\%)$. For the stronger ConvLSTM model (\Cref{fig:tempb}), the best results are obtained for $8$ cell state metrics, that is, $R^2 =63.33\%(\pm0.59\%)$, whereas the baseline metrics only obtain $R^2$ values up to $62.76\%(\pm0.58\%)$. These results are in line with the findings in \cite{Rottmann.2020} and \cite{Maag.2020}, that is, stronger segmentation models yield worse meta performance with respect to $R^2$. Moreover, the analysis of time series metrics in \cite{Maag.2020} showed a performance gain for linear models, whereas, the stronger gradient boosting models do not benefit as much from time series metrics. We observe the same effects with regard to temporal metrics. Finally, with regard to meta classification based on the strong model (\Cref{fig:tempc}), we observe that all models benefit from the temporal metrics, while the linear model outperforms the shallow LSTM and neural network by $0.15pp$ and $0.26pp$, respectively. Note that, even though the linear model is only slightly better than the shallow network, this result is not in line with the findings of \cite{Rottmann.2020} and \cite{Maag.2020}, where the neural networks outperformed the linear models in all experiments.

Next, we investigate time series metrics $\{V_{m,t},V_{m,t-1},\dots,V_{m,t-T}\}$ with\break $m=9$ and a length of up to $T=10$ previous timesteps, yielding $11$ different sets of metrics. The results are summarized in \Cref{tab:timeseriesmetrics}. Since the gradient boosting model performs best in \cite{Maag.2020} as well as in most of our experiments, we consider the gradient boosting model equipped with the metric set $\{U_t, U_{t-1}, \dots, U_{t-10}\}$ as the baseline model. This baseline is outperformed for both meta tasks and both, the strong and the weak model. For the weak model, we achieve classification accuracy up to $95.25\%(\pm0.22\%)$ with our proposed LSTM meta model considering $1$ cell state metric. For meta regression, we obtain $R^2$ up to $74.31\%(\pm0.33\%)$ by the gradient boosting model. For the strong model, we achieve best results for the classification task by means of the gradient boosting model, whereas our proposed LSTM meta model outperforms the gradient boosting model in the regression task yielding $R^2$ values of $69.00\%(\pm0.98\%)$ with $6$ considered cell state metrics.

\begin{table}[t]
	\caption{Results for meta classification and regression based on temporal metrics for different meta models and the GB baseline from \cite{Maag.2020} for both, the weak (W) and the strong (S) model. The super script denotes the number of frames, where the best performance and in particular the given values are reached. The best results are highlighted.}\label{tab:timeseriesmetrics}
	\begin{center}
		\scriptsize
		\begin{tabular}{|| c || c || c | c | c | c ||} 
			\hline
			\multicolumn{6}{||c||}{Meta Classification $IoU_{adj}=0,>0$} \\
			\hline
			\multicolumn{6}{||c||}{Baseline \cite{Maag.2020} (W): $ACC = 94.93\%(\pm0.32\%)$ $AUROC = 94.99\%(\pm0.35\%)$} \\
			\multicolumn{6}{||c||}{Baseline \cite{Maag.2020} (S): $ACC = 96.03\%(\pm0.18\%)$ $AUROC = 94.12\%(\pm0.43\%)$} \\
			\hline
			\multicolumn{2}{||c}{} &
			\multicolumn{1}{c|}{GB} &
			\multicolumn{1}{|c|}{LR} &
			\multicolumn{1}{|c|}{LSTM} &
			\multicolumn{1}{|c||}{NN} \\
			\hline
			\multirow{2}*{$ACC$}
			& W & $94.95\%(\pm0.24\%)^9$ & $94.64\%(\pm0.24\%)^8$ & \bm{$95.25\%(\pm0.22\%)^1$} & $94.09\%(\pm0.23\%)^6$ \\
			& S &  \bm{$96.15\%(\pm0.17\%)^1$} & $95.88\%(\pm0.23\%)^1$ &  \bm{$96.15\%(\pm0.17\%)^9$} & $95.54\%(\pm0.30\%)^1$ \\
			\hline
			\multirow{2}*{$AUROC$}
			& W & $95.00\%(\pm0.28\%)^1$ & $94.24\%(\pm0.34\%)^1$ & \bm{$95.04\%(\pm0.22\%)^1$} & $93.32\%(\pm0.47\%)^1$ \\ 
			& S &  \bm{$94.23\%(\pm0.42\%)^9$} & $92.85\%(\pm0.39\%)^1$ & $93.65\%(\pm0.46\%)^1$ & $91.92\%(\pm0.64\%)^0$ \\
			\hline
			\multicolumn{6}{||c||}{Meta Regression $IoU_{adj}$} \\
			\hline
			\multicolumn{6}{||c||}{Baseline \cite{Maag.2020} (W): $\sigma = 0.153(\pm0.002)$ $R^2 = 74.00\%(\pm0.65\%)$} \\
			\multicolumn{6}{||c||}{Baseline \cite{Maag.2020} (S): $\sigma = 0.161(\pm0.001)$ $R^2 = 68.27\%(\pm0.53\%)$} \\
			\hline
			\multicolumn{2}{||c}{} &
			\multicolumn{1}{c|}{GB} &
			\multicolumn{1}{|c|}{LR} &
			\multicolumn{1}{|c|}{LSTM} &
			\multicolumn{1}{|c||}{NN} \\
			\hline
			\multirow{2}*{$\sigma$}
			& W & \bm{$0.154\%(\pm0.001\%)^6$} & $0.168\%(\pm0.001\%)^6$ & $0.157\%(\pm0.003\%)^6$ & $0.157\%(\pm0.003\%)^7$ \\ 
			& S &  \bm{$0.162\%(\pm0.002\%)^8$} & $0.173\%(\pm0.002\%)^8$ &  \bm{$0.162\%(\pm0.002\%)^8$} & $0.163\%(\pm0.002\%)^8$ \\
			\hline
			\multirow{2}*{$R^2$}
			& W & \bm{$74.31\%(\pm0.33\%)^0$} & $69.15\%(\pm0.46\%)^3$ & $73.58\%(\pm0.74\%)^3$ & $73.54\%(\pm0.39\%)^0$ \\
			& S & $68.97\%(\pm0.81\%)^1$ & $64.44\%(\pm0.51\%)^1$ &  \bm{$69.00\%(\pm0.98\%)^6$} & $68.53\%(\pm1.04\%)^6$ \\
			\hline
		\end{tabular}
	\end{center}
\end{table}

\section{Conclusion and Outlook}
In this paper, we extended the approach from \cite{Rottmann.2020} and \cite{Maag.2020} for deep ConvLSTM networks. We introduced temporal metrics based on the cell states broadcasted through LSTM cells as additional inputs for meta classification and regression. In our experiments, we studied the influence of different numbers of considered cell state metrics for four meta models, i.e., linear models, gradient boosting, shallow neural networks as well as shallow LSTM models. Moreover, we investigated the influence of LSTM meta models for time series metrics proposed in \cite{Maag.2020}. In all experiments, our approach slightly improved the state of the art results \cite{Rottmann.2020} and \cite{Maag.2020}. More precisely, we achieve classification accuracy of $96.15\%(\pm0.17\%)$ and $AUROC$ of $95.04\%$ $(\pm0.22\%)$ using our proposed LSTM meta model with temporal metrics. For meta regression, we obtain $R^2$ values of $74.31\%(\pm0.33\%)$. We plan to develop further LSTM-based metrics for uncertainty quantification applied to the task of predicting several steps into the future.

\subsubsection{Disclaimer} The results, opinions and conclusions expressed in this publication are not necessarily those of Volkswagen Aktiengesellschaft.

\bibliographystyle{splncs04}
\bibliography{TemporalPerformancePredictionforDeepConvLSTMs}

\end{document}